# Voice Over Body? Older Adults' Reactions to Robot and Voice Assistant Facilitators of Group Conversation

Katie Seaborn[1] · Takuya Sekiguchi[2] · Seiki Tokunaga[2] · Norihisa P. Miyake[2] · Mihoko Otake-Matsuura[2]

## Abstract
Intelligent agents have great potential as facilitators of group conversation among older adults. However, little is known about how to design agents for this purpose and user group, especially in terms of agent embodiment. To this end, we conducted a mixed methods study of older adults' reactions to voice and body in a group conversation facilitation agent. Two agent forms with the same underlying artificial intelligence (AI) and voice system were compared: a humanoid robot and a voice assistant. One preliminary study (total n = 24) and one experimental study comparing voice and body morphologies (n = 36) were conducted with older adults and an experienced human facilitator. Findings revealed that the artificiality of the agent, regardless of its form, was beneficial for the socially uncomfortable task of conversation facilitation. Even so, talkative personality types had a poorer experience with the "bodied" robot version. Design implications and supplementary reactions, especially to agent voice, are also discussed.

## 1 Introduction

Intelligent agents are gaining momentum as a tool for communication and social interaction in older adults' daily lives. Social robots, voice assistants, smart devices in the home, and voice user interfaces (VUIs) on an array of devices are increasingly being explored with older adults [1–9]. Virtual assistants like Amazon's Alexa (via the Echo smart speaker) and Apple's Siri (via the iPhone and other MacOS-based devices) are common examples. Being widespread on consumer devices, they offer new and potentially more natural forms of human-computer interaction to older adults [3, 7, 10–13]. Regardless of the "body," these voice-based agents tend to rely on speech-based interaction in query-style and conversation-like formats [12]. Additionally, multiple users are often supported or supportable in theory, leaving room for social and collaborative activities and effects [14–16].

Social applications may be particularly relevant for older adults, who are at risk of isolation and the severe negative impacts it can have on well-being and physical health [17, 18]. Activities like agent-facilitated group conversation may enhance older adults' social lives as well as provide cognitive benefits and entertainment [19, 20]. For example, agent facilitators could be deployed in group homes, community centres, hospitals and care facilities, or even online through a video conferencing medium like Zoom to enable and manage social networking among older adults. Time of day, hours of work, and exhaustion would not be factors with an agent facilitator. Such an agent facilitator could provide conversational scaffolds for levelling the playing field among older adults of varying cognitive abilities, ensure that everyone has a say and that no one talks too much, offer prompts during lulls in conversation, and more. At present, the possibilities are vast and mostly uncharted for older adults.

We also know little about the fit between voice-only, "bodyless" intelligent agents and older people [3, 21–23]. Older adults have long been considered a user group affected by a "digital divide" [24–26]. This "gap" is closing as technology becomes ubiquitous in daily life, but is compounded when new form factors and interaction paradigms emerge, such as VUIs [6]. One survey found that only

✉ Katie Seaborn
seaborn.k.aa@m.titech.ac.jp

[1] Department Industrial Engineering and Economics, Tokyo Institute of Technology, Tokyo, Japan

[2] RIKEN Center for Advanced Intelligence Project (AIP), Tokyo, Japan

1% of research on voice in human-agent interaction (HAI) included older adults [12]. When it comes to voice-based intelligent agents, much work is needed to uncover which embodiments are ideal in terms of form factor, interaction style, relationship to the user, situatedness in the context of use and environment, and more. Visible and tangible agent forms that offer a clear source of the voice and "mind" of the agent, such as robots and virtual characters, may be more familiar and easier to understand for older adults. At the same time, voice-only agents and interfaces that do not require the use of eyes or hands-on interaction may be more usable and accessible than "bodied" options. Voice interaction may support older adults experiencing age-related reductions in vision and motor ability in ways that visual and tangible interfaces cannot [27]. Additionally, the displays and controls older adults tend to find difficult to use are largely absent in voice-only systems [11]. Finally, the cost and resources required to create and deliver "bodied" agents, such as robots, is higher than "bodiless" voice assistants, so there are practical implications. In short, the merits and demerits of these form factors for older adults need to be teased out and explored. We cannot assume that findings from other age groups will translate to older people. As a matter of inclusion, research that targets older adults as a user group is needed.

To this end, we explored the question of "voice or body" by evaluating the role of agent embodiment in older adults' experiences of group conversation facilitated by a voice-based intelligent agent. We compared two "bodies" of the agent, which used the same underlying artificial intelligence (AI) and voice system: a small humanoid robot and a speaker-based virtual assistant. We asked: (RQ1) *Do older adults react differently to voice and robot embodiments of an intelligent conversation facilitation agent?* and (RQ2) *If the "body" matters, what specific features are salient?* One preliminary study was run with older adults. The main objective was to validate a new approach to assessing agent embodiment through a questionnaire designed to target specific features of the robot's body. A second objective was to access early insights, especially about the common factor between the agent forms: the voice. In the main study, a new cohort of older adults was exposed to both forms of the agent over four sessions. The main objective was to compare agent voice and body within the group conversation context. A mixed methods experimental design approach [28] was taken to capture a well-rounded view of older adults' experiences, attitudes, and behaviours. This involved open and closed questionnaire responses, speech data recorded by the system, observation notes from the sessions and group retrospective think aloud protocols at the end of each day, and a follow-up interview with an experienced human facilitator.

We offer four contributions. First, we provide findings on older adults' experiences with voice-based intelligent agents employed as facilitators of group conversation. Specifically, we provide comparative findings on robot and voice morphologies ("bodies") through a controlled experimental design supplemented by qualitative insights. Second, we offer a methodology and research tool—an agent embodiment questionnaire—for evaluating specific features of agent "bodies," which complements existing high-level agent embodiment instruments. Third, we contribute evidence showing that individual factors in group conversation settings relate to agent embodiment. Specifically, our findings suggest that highly talkative older adults may be better suited to voice-only agents. Fourth, we articulate these findings in a set of design implications. This work builds on the agent embodiment literature for "voice" and "body." It also extends this body of work to conversation contexts and older adults, a much-needed trajectory in a booming area of study with an understudied cohort.

## 2 Theoretical Background

We start by reviewing the work on intelligent agents and embodiment generally. From this, we develop a hypothetical framework guided by our two RQs. Then, we motivate our hypotheses. We start by considering the small corpus of existing work involving older adults. We then turn to individual and social factors in group conversation with people that may extend to situations involving intelligent agents.

### 2.1 Intelligent Agents and Embodiment

Research on agent embodiment has a long and rich history. *Embodiment* refers to an agent's morphology—its physical form factor, including sensors and actuators—as well as how it interacts with its environment and other agents, typically humans, when situated within a given context [21, 23]. In robots, embodiment is typically expressed through visible, tangible form factors supplemented by verbal and nonverbal behaviours. The Nao platform, for instance, is made up of a physical robot that can move through space, make gestures, listen and speak, and detect objects and agents in the environment through computer vision. The physical version of Nao also has a virtual complement that can be instantiated in a 3D computer simulation, typically used for testing out behaviours before deployment in the real world. In HAI research, embodiment has primarily been approached through a focus on morphology. Indeed, the growing prevalence of voice assistants have re-sparked interest in the "body" question for intelligent agents. Some have argued that a perceivable body is necessary (e.g., [29]),

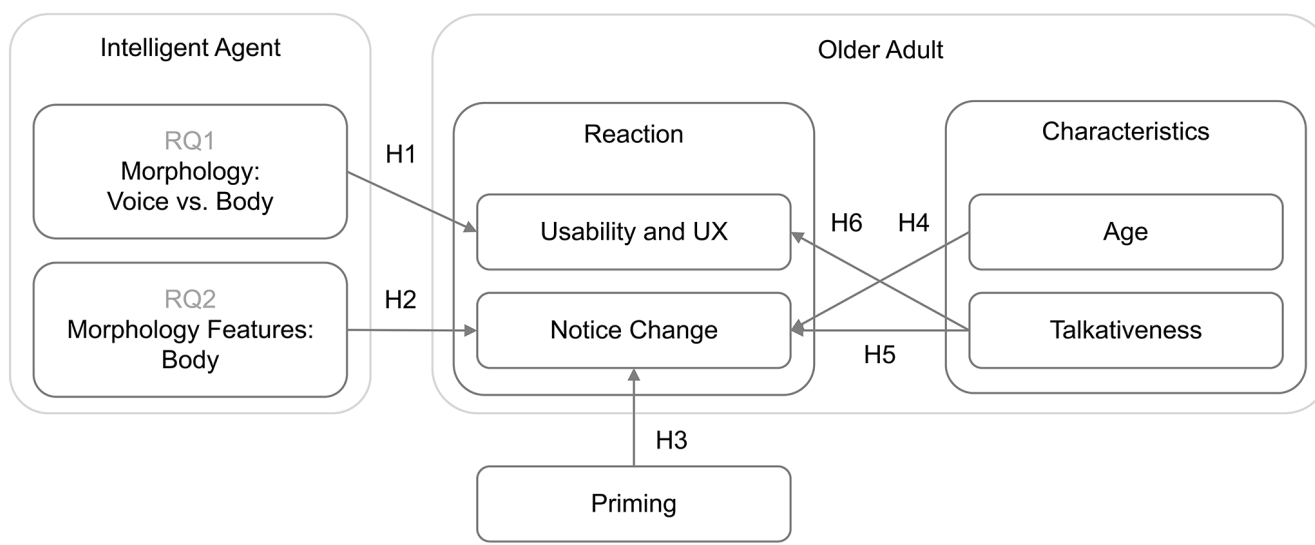

**Fig. 1** Hypothetical framework situating the RQs and hypotheses with respect to the intelligent agent and its embodiments, researcher priming, and the older adult participants. Arrows indicate theoretical direction of the effect, e.g., age is hypothesized to influence ability to notice a change

while others have suggested that it depends on the experiential measure, activity, and/or context of use (e.g., [16, 30–34]). Practically, voice-only and "bodied" form factors each have their positives and negatives. Robots, for instance, take a long time to develop, requiring expertise to finesse and special resources to produce and distribute. While portable, robots need to be shipped and involve setup and user training. Voice assistants, in contrast, only need a speaker to be instantiated. Given the cost and availability of speaker systems, as well as the ubiquity of speakers in personal devices like smartphones and laptops, such form factors are comparatively cheap and portable. These practical concerns need to be balanced against factors of user experience (UX), usability, and appropriateness in the use case and for the user group.

Some work has provided evidence that "bodied" agents are superior to "bodiless" ones in certain ways. Shamekhi et al. [16] compared voice-only and face avatar versions of a conversational agent, finding that giving the agent a face improved its reception socially, with less clear effects on task performance. Luria, Hoffman, and Zuckerman [35] found that robotic control interfaces for smart homes were superior to voice-based interfaces in terms of feelings of control and situation awareness. Kontogiorgos et al. [36] found that a social robot garnered better engagement and sociability ratings compared to a smart speaker. So far, there is little consensus on when a body is needed and, if so, what features are important, for whom, and under what conditions [21, 23]. Additionally, most work has focused on high-level comparisons (e.g., avatar versus voice-only) and measures of preference, trust, and task performance.

At present, it is difficult to determine which intelligent agent embodiments are best and what features of those morphologies are most important. As such, we have chosen two intertwined RQs that aim to clarify (RQ1) whether the body matters and (RQ2) if so, what specifically matters. Additionally, in the event that body matters (RQ2), we have designed an agent embodiment questionnaire to pinpoint salient features of the robot's morphology. We have employed this tool in a comparative research design to test whether these features are perceivable by and meaningful to participants, thus providing a level of detail not typically evaluated. Figure 1 represents our hypothetical framework guided by these RQs, our research design, and our literature review. We motivate the hypotheses represented in this figure next.

### 2.2 Older Adults and Intelligent Agent Embodiment

While agent embodiment in general needs greater attention, it is especially true for older adults as a user group. Some work suggests that suitability of agent embodiment differs based on context of use, the older adult's life circumstances, and the agent's relationship to those circumstances, i.e., medical care versus fitness, companion versus assistant, in home versus in hospital, etc. [37–39]. Most of the work so far, however, has focused on the *morphology* of the agent. In particular, the greater portion of the work until recently has centered on agents with a visible body [3, 12]. Most of these feature humanoid or zoomorphic forms of morphology. Paro the seal [40, 41], Aibo the robotic dog [42, 43], and the Huggable with its teddy bear form factor [44] are well-known examples. Alternatives are starting to be explored, including object morphologies, such as Hugvie the pillow [45], and abstract geometric robots [39]. As yet, "bodiless" morphologies that tap into other sensory modalities, such as voice and speech, remain underexplored for older adults.

Centering older adults as users of voice-based and voice-only agents, interfaces, and systems—voice assistants, conversational user interfaces (CUIs), VUIs, and smart speakers, to name a few—is nascent but needs more attention [3, 7, 10, 27, 46, 47]. Most work has focused on commercial examples available in smartphones and smart speakers, notably Apple's Siri [3, 7], Amazon's Alexa [1, 7], and Google Home [11]. Some recent work has started to address this gap. For instance, in a preliminary study on within the context of healthcare, Sin and Munteanu [6] found that the form of the agent was connected to the task, with voice agents preferred for generic information finding and humans preferred for personal care. Trajkova and Martin-Hammond [7] focused on older adults' use of Amazon's Alexa (via the Echo smart speaker) over a year period. They found that attitudes and behaviour, particularly abandonment, were mediated by meaningful cases, shared spaces, and perceived lack of ability. While this work represents a step in the right direction, it takes for granted that voice UX is ideal.

So far, there is little comparative work on agent embodiment and older user groups. Most studies feature one type of agent, and usually this agent is a social or humanoid robot (e.g., [48]) or a human (e.g., [20]). The lack of comparative data makes it difficult to conclude what type of agent is ideal for what situation. We sought to remedy this by combining a qualitative approach inspired by previous work with a comparative experimental design focusing on a key factor—agent embodiment. In line with most forms of agent embodiments in work involving older adults, past and present, we compared humanoid robot and voice assistant embodiments.

As discussed, scholars (e.g., [29]) have argued that a perceivable body offers better UX in some way, and there is some research to support this (e.g., [16]). What little work that exists on older adults tends to support this, as well. Matsuyama et al. [49] developed and tested a framework for robotic facilitation, relying on participants' utterances to guide the robot's behaviour and prompts. They found that relying on paralinguistic information was insufficient, proposing that parsing visual information, such as eye gaze, may be necessary. Given the results so far, we would expect that a robot would outperform a voice assistant, but this has not been explored. This leads to the following hypothesis for our comparative study:

*H1: The agent's robot embodiment will have higher usability and UX scores than its voice form.*

If the body is important, then people will surely pay attention to it in some way. Indeed, some work on eye gaze duration found that this was true for robotic embodiments over bodiless, voice-based ones (e.g., [36]). What we do not understand well is what, if any, specific visual features of the body people are attending to, and why. Indeed, it is possible that people are instead taking in the gestalt: the whole rather than the parts. Even so, it would be useful for designers and roboticists to know if there are certain features that are important to include (or exclude). For instance, people are naturally predisposed to seeing faces and even emotional expressions in objects, a phenomenon called pareidolia [50]. In effect, we are primed to see human qualities even when the barest hint of a cue exists. Therefore, a static facial expression, such as a peaceful smile, may significantly impact a robot's reception and maintain the tone over the course of an interaction. As such, our robot was designed with a fixed expression.

At the same time, we must somehow isolate the effect, if any, of certain features. One way to do this is to change the feature and re-evaluate the measures afterwards using a repeated measures or crossover design [51]. Yet, a wealth of research has shown that people are susceptible to change blindness in visual information, where we can fail to notice certain visual changes, even large ones [52]. This can be avoided by directing people's focus (or locus) of attention. For example, Bae and Kim [53] found that people were better at change detection in animated robots compared to inanimate ones. Likewise, a conversation facilitator, even one with a static facial expression, may bring attention to its visual form when calling on certain participants. For the robot version of our agent facilitator, we can subsequently hypothesize the following:

*H2: Most participants will notice a change to the robot body.*

Explicitly pointing to certain features in advance, or priming the locus of attention, can also reduce change blindness (e.g., [54]). Priming may then be a way of evaluating the degree to which noticing a visual change is due to the robot's natural embodiment. As such, we hypothesize the following as well:

*H3: Priming will increase participants' ability to notice a change to the robot body.*

Age is known to play a role in older adults' cognitive performance [55], including with respect to change detection [56]. As we age, we tend to experience declines in various cognitive functions. It is therefore possible that older adults in different age cohorts—young-old, old-old, and oldest-old [57]—may perform differently. Discovering whether this is the case is important in establishing the reliability of agent

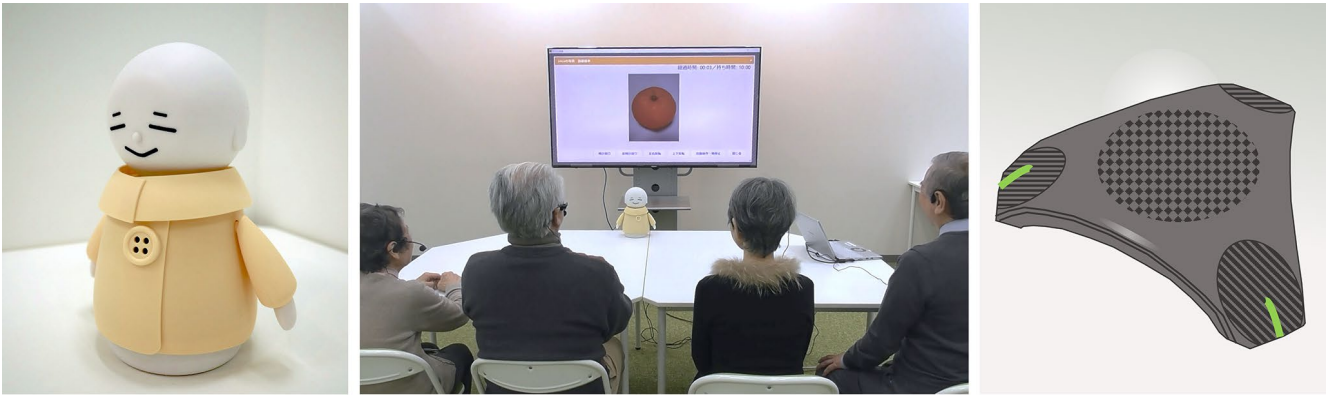

**Fig. 2** Bono the robot or "Bono Bot" (left), the setting for the study (middle), and the voice assistant version or "Bono Voice" with a smart speaker form (right)

embodiment research tools for older adult user groups. As such, we hypothesize the following:

> H4: Age will affect participants' ability to notice a change to the robot body.

### 2.3 Individual and Social Factors in Group Conversation

In group conversation contexts involving people, individual and social dimensions come into play. In particular, *dominance* as expressed through speaking behaviour and other social and behavioural cues (such as role in the group) is a well-established factor [58]. In their meta-analysis, Mass [58] showed that speaking time or *talkativeness* is significantly associated with dominance. In other words, a person's talkativeness can affect how much or how little they participate in group conversation, and on the flip side, how much or how little their behaviour influences or is influenced by other people's talkativeness, as well. There are many ways to measure talkativeness. While general measures of personality, especially extraversion, can be used, these do not pinpoint talkativeness. Some work going back decades has questioned the link between extraversion and talkativeness, In a classic study, Thorne [59] found that conversational style was a more significant factor in distinguishing pairs of extraverts and introverts. McLean and Pasupathi [60] captured extraversion through the Big Five scale as well as a custom subjective talkativeness measure. They found that when removing the talkativeness item from the extraversion scale, the results were the same, indicating that talkativeness was a distinct factor. In our study, the agent records and calculates the speaking time for all participants, in line with objective measures. We also propose a short subjective self-report measure for targeting talkativeness, detailed in 5.4.1.

Group conversation is a distracting activity that reduces people's ability to attend to other matters, such as changes in the landscape while driving [61]. If this is true, we may expect that participants who talk a lot may be so engaged in expressing their point of view that they do not attend to others, including an intelligent agent facilitator. In our case, we would expect that people would not pay attention to and/or remember the robot's physical features, including any changes made to those features. We can operationalize this as a factor of how many key features of the (robot's) body a person can recall with accuracy: an *awareness-morphology* score. We thus hypothesized:

> H5: High talkativeness will lead to lower awareness-morphology scores for the robot version of the agent.

Role within the group is also associated with dominance and talkativeness [58]. Roles may be interpreted as having certain functions and power, creating a hierarchy of dominance, even a temporary one. A talkative participant may then take issue with a strict facilitator, human or otherwise. In our case, the intelligent agent was designed to monitor and intervene when people talk too much or too little, just as an experienced human facilitator would do. We would thus expect participants with high talkativeness scores to have worse UX with the agent. We hypothesized:

> H6: High talkativeness will lead to worse UX scores for both agent forms.

Next, we introduce the system, including the intelligent agent and its embodiments.

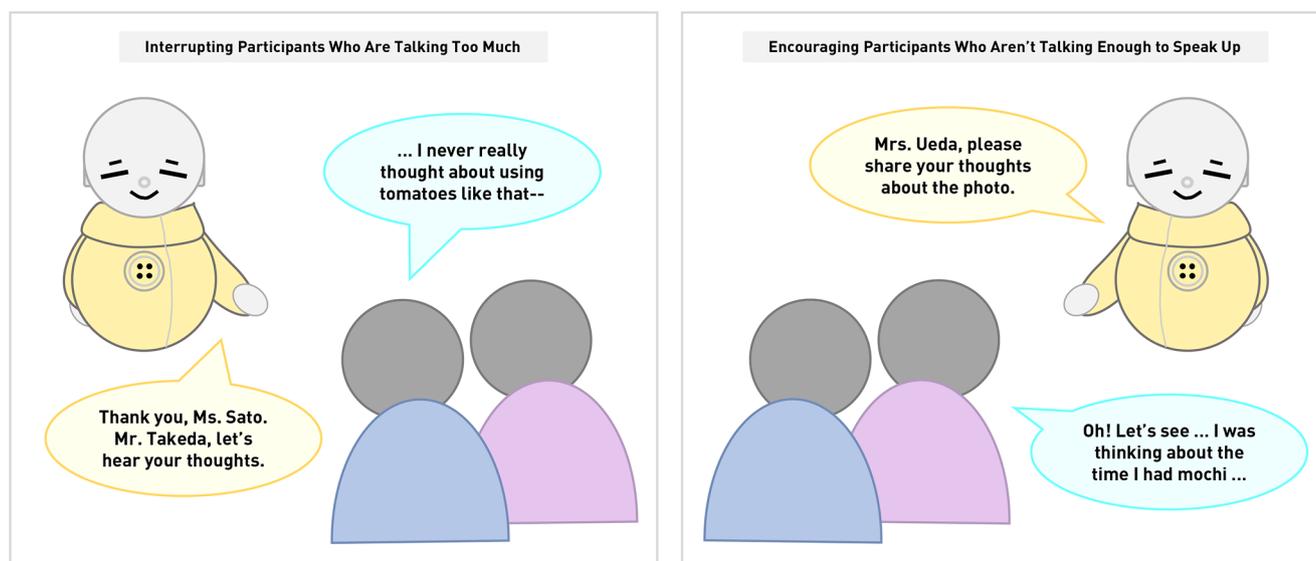

**Fig. 3** Demonstration of how Bono facilitates group conversation. Left: Bono interrupts a talkative participant. Right: Bono prompts a quiet participant to contribute

## 3 System

We used two forms of an intelligent agent for facilitating group conversation: a humanoid robot, representing the "bodied" form factor (Fig. 2), and a voice assistant, representing the "voice-only" or "bodiless" form factor. Both forms used the same conversation-parsing AI, which was linked to a particular group conversation protocol designed for older adults called the Coimagination Method. We describe in detail below.

### 3.1 Group Conversation Activity: Coimagination

The Coimagination Method is a small group (2–6 members) conversation protocol that aims to mitigate the risk of cognitive decline in older adults [62]. As a media-based method, it uses photos as prompts to spark memories and discussion of experiences [63]. We chose Coimagination because it is a type of talk-focused reminiscence therapy for older adults [63–68]. Benefits include feelings of ownership, reduced barriers with visual media, a social experience, the potential for building and maintaining social networks, and mitigating risk factors related to cognitive decline [19, 67]. Moreover, Coimagination, unlike other protocols, came with a speech-interpreting AI platform that we could use as the baseline intelligent agent. More details about the steps and tasks involved in Coimagination are detailed in the procedure section below.

### 3.2 Intelligent Agent and Its Forms: Bono Bot and Bono Voice

"Bono" is an AI-based intelligent agent designed to facilitate small group conversation according to the Coimagination protocol. The AI powering Bono records, parses, and responds to conversation from multiple participants in real-time. In this way, it can control the amount of "airtime" a given participant has, such as by encouraging certain people to speak up (Fig. 3, middle) or allow others to speak (Fig. 3, left). Bono's voice was Hikari, a synthetic feminine Japanese voice generated with the ReadSpeaker speechMaker system[1]. We chose this voice because it is gender-ambiguous and has an upbeat, calm vocal affect that we expected to soften the strictness of the agent's facilitation. Bono's body (Fig. 2, left) was modelled after Japanese Kokeshi dolls, a type of woodcraft object from the Edo period. It has been iteratively developed over several years in collaboration with engineering students, older adults who have participated in the Coimagination Method, and a local robotics company. It can move in response to the people around it by turning its head to look at a particular person or moving its arms to provide simple gestures. The smart speaker version, or "Bono Voice," can be instantiated through any speaker system, including smartphones. For this study, we used a used a triangular teleconference speaker (Fig. 2, right). Both forms of Bono, via this shared AI, can sync with the photo panel system used to facilitate the Coimagination Method. Refer to Online Resource 7 for a video of Bono's actions when carrying out a session.

---

[1] JP: https://readspeaker.jp EN: https://www.readspeaker.com.

### 3.3 Supporting System: Photo Display Panel

The photo display panel (hereafter "panel") is used to present the photos taken by participants in a highly controlled and timed fashion (Fig. 2, right). It can be projected or shown via a monitor. The main elements are the photos and a timer. It also has an admin interface for setting up sessions and syncing with Bono. When used with Bono, start and end times for speaking are verbalized by Bono; without Bono, a human facilitator is expected to verbalize this feedback.

## 4 Preliminary Study

We conducted a preliminary study before the main study. Our main purpose was to test and refine the agent embodiment questionnaire and especially the awareness-morphology measure. We also wished to test the usability and UX instruments, as well as collect initial reactions to Bono's voice in preparation for using the voice assistant in the main study. A list of all measures and instruments used across studies is available in Online Resource 5. We can share the data set upon reasonable request, i.e., for meta-analysis purposes.

### 4.1 Participants and Setting

Older adults (10 men and 14 women; 24 total in 6 groups) were recruited from in and around the city using a third-party recruiter. Ages ranged from 65 to 78 (M = 72.3, SD = 4.1, MD = 71, IQR = 6.5). All had no or mild cognitive impairment, basic technology skills (e.g., smartphone use), and were able to travel to the study location. Participants were mid-way through a 12-week course on the Coimagination Method. The study was conducted in a discussion space with seats, a central table, and a large screen. Ethics approval was obtained from the RIKEN research ethics committee under No. Wako3 30-11(3). Informed consent was obtained from all participants and compensation was ~ $12 US per hour plus transportation.

### 4.2 Procedure

Each trial was comprised of a 20-25-minute Coimagination session. At least a week before the session, participants were given a theme that they had to take a photo about, e.g., "A seasonal food." Before the session, participants transferred the photo to the technical staff, who uploaded it to the panel. The session began with an introduction by an assistant about the Coimagination Method and Bono, and then facilitation was passed to Bono. Next, each participant was asked by Bono to explain and reminisce about the photo that they took in 1 min, in detail, using all of the time allotted. Bono managed the time and moving from participant to participant. Then, Bono gave each participant two minutes to respond to any questions or comments (i.e., Q&A) from other participants about their photo. Bono facilitated by interrupting those who were speaking too much and calling on those who were speaking too little. After the session, participants filled out the agent embodiment questionnaire individually. They were then given the opportunity to debrief and ask questions. Compensation was processed by the recruitment agency later on. The trial then ended.

### 4.3 Instrument and Measures

The instrument was the agent embodiment questionnaire we created. It was comprised of a series of closed and open-ended items for the awareness-morphology measure, as well as usability and UX scales. All were subjective self-report measures and most used a 5-point Likert scale, unless stated otherwise. See Online Resources 3 and 4 for the English and Japanese versions of the questionnaire, respectively.

#### 4.3.1 Awareness-Morphology

We developed a measure of awareness of the robot's morphology involving accuracy and self-confidence, inspired by Bornstein and Zickafoose [69]. As a measure of accuracy, two items asked about the colour of the robot's face and body from eleven options (including "I do not know"). A third item asked respondents to indicate the robot's (unchanging) facial expression out of seven options (one was "I do not know"). Self-confidence was assessed by directly asking participants how confident they were with the previous answer. A 4-point confidence scale was used, e.g., "I am very confident."

#### 4.3.2 Usability and UX

We used 16 items inspired by the applied Usability, Social Acceptance, User Experience, and Societal Impact (USUS) framework for HRI [70]. We adapted 3 items from the Japanese version of the System Usability Scale (SUS) [71, 72] for the HRI context, specifically: "I think I would use this robot in other group conversation situations," "The functions of this robot are well-organized; for example, how it moved, how it facilitated, etc.," and "The robot was not effective at facilitating." We created a custom scale to target key features of the group conversation and agent facilitation context. Example questions include: "The agent's facilitation was easy to understand," "I liked the appearance of the agent," and "The agent is better than a person at interrupting speech." For the voice, we asked respondents to "Choose three words that describe your impression of the voice and its characteristics."

### 4.4 Data Analysis

Descriptive and inferential statistics were used to evaluate individuals and groups across the measures derived from the questionnaire responses. For the awareness-morphology measure, a product of the correct scores and self-confidence was created for inferential analyses. Kendall's tau-b correlations were used to compare correct answers with subjective confidence, and these were then multiplied to create a product variable for use in other analyses. An overall measure of usability/UX was created by summing the sums of each scale, in line with treating Likert scale data [73]. Individual and group level measures were generated. All measures used in analyses had an internal consistency of Cronbach's alpha $\alpha > 0.90$ (see Online Resource 6 for details). Pearson correlations were used to indicate relationships between measures of interest. Qualitative responses were translated from Japanese by the PI and double-checked by another author fluent in Japanese. Content analysis of this data was conducted with a focus on frequency of codes [74].

### 4.5 Results

Descriptive statistics can be found in Table 1 in Online Resource 6.

#### 4.5.1 Awareness-Morphology

We found that 16 older adults out of 24 (67%) selected the correct face colour, 12 (50%) selected the correct body colour, and 20 (83%) selected the correct facial expression. Six people (25%) scored perfect. In terms of confidence, 13 (54%) had high confidence, 10 had low confidence (42%), and one abstained. Correlations were not found between confidence and correctness scores, $\tau b = -0.080, p = .67$. This means that confidence did not affect accuracy.

#### 4.5.2 Usability and UX

Pearson correlations indicated that the usability and UX measures were correlated. A high positive correlation between the Bono-UX and USUS scores, $r(22) = 0.825$, $p = .008$, $R^2 = 0.681$ was found. Moderate positive correlations between the Bono-UX and SUS scores, $r(22) = 0.685$, $p < .05$, $R^2 = 0.470$, and between the USUS and SUS scores, $r(22) = 0.701$, $p < .05$, $R^2 = 0.492$, were found. Based on this, we summed the measures together to create one UX construct. A moderate correlation was found between this construct and satisfaction with the robot's voice, $r(22) = 0.726$, $p < .05$, $R^2 = 0.528$.

Most (67%) somewhat liked Bono's voice (16/24). 37% felt that the voice had an accent. Another 37% felt that the voice was pleasant. 22% felt that the voice was mechanical, while 19% felt that it was monotonous (together, 41%). 11% wrote that the voice was high-pitched. 11% assigned a gender to the voice, two writing masculine and one writing feminine. When asked about how to improve Bono's voice, half of comments indicated that older adults did not want the voice to change (13/27 comments, or 48%). The top request was enabling greater variety in what Bono said (6 requests) and including praise (3 requests). A small negative correlation was found between voice satisfaction and assigning the "mechanical" characteristic to the voice, $r(24) = -0.615$, $p < .01$, $R^2 = 0.378$. No other characteristics were correlated.

### 4.6 Findings & Discussion

The basic visual features of the robot's morphology varied in saliency from participant to participant. About half correctly assigned face and body colours as well as facial expression. Individual differences may be explained by each participant's relative experience with the robot-directed version of the Coimagination Method. But they may also be explained by other factors, particularly level of distraction in the conversation task, i.e., one's talkativeness. We sought to confirm this in the main study.

The lack of relationship between confidence and accuracy scores tells us that we cannot rely on subjective measures alone. This is likely due to confidence effects, where people rate themselves as over- or under-confident depending on the perceived difficulty of the task [75]. The instrument itself caused no issue with the older adults, who were able to understand and complete it reasonably quickly.

We were also able to confirm the complementarity, feasibility, and usefulness of the added usability and UX measures as well as gather initial reactions to Bono's voice. Older adults were satisfied with voice. Even so, the accent and perceived mechanical quality of the voice may have affected satisfaction ratings for some, i.e., people who did not like the voice did not like it because it was mechanical. Participants' top request was for greater variety in what the robot said, but because speech content is strictly tied to the conversation method, we did not make any changes.

## 5 Main Study

The goal of this study was to directly compare robot and voice assistant embodiments for intelligent agent conversation facilitation with older adults. To this end, we used a quantitatively-driven [76] embedded design [28], with qualitative methods added to a controlled experimental design. Such a combined approach can provide a more well-rounded answer to research questions than quantitative or qualitative approaches alone, as well as help explain quantitative results. In our case, we used a combination of hypothesis testing with inferential statistics and qualitative analysis of questionnaire responses, speech data, participant reactions and commentary from a group retrospective think-aloud protocol (RTA), the PI's observation notes, and data from an interview with an experienced human facilitator who was familiar with Bono Bot. We can share the data set upon reasonable request, i.e., for meta-analysis purposes.

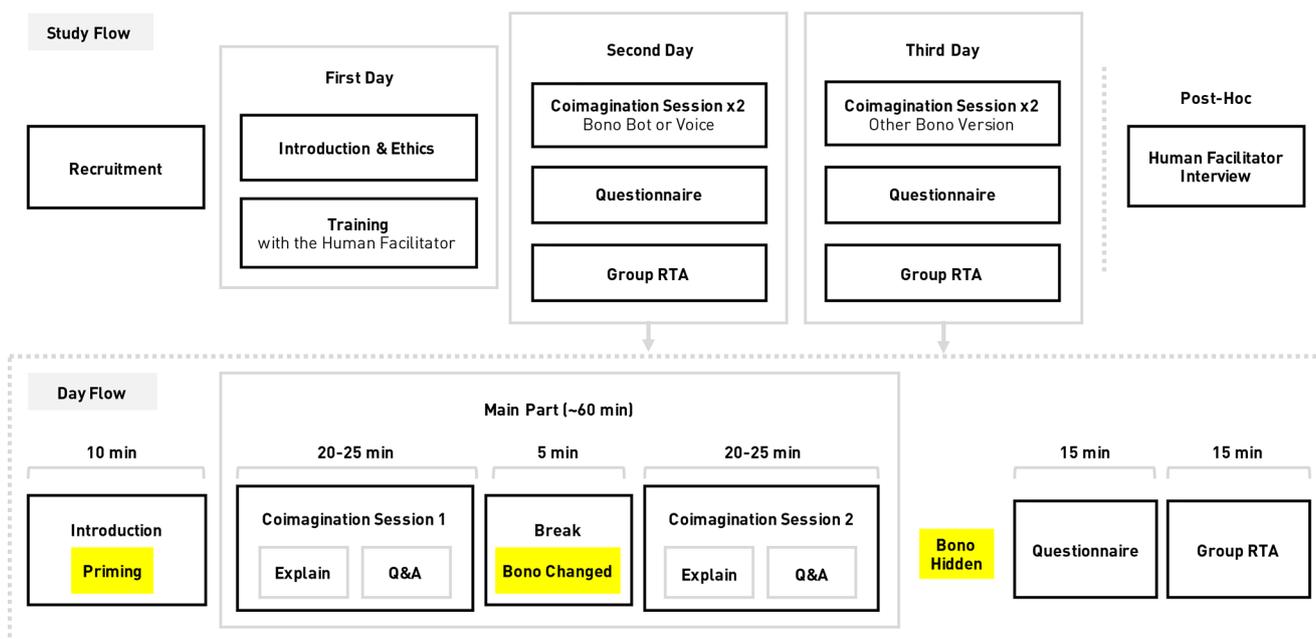

**Fig. 4** Overall study flow per group (top) and per day (bottom). The parts in yellow are only applicable to the robot condition

### 5.1 Experimental Design

We used a $2\times2\times2$ mixed factorial design with the within-subjects factor *agent form* (robot and voice assistant), the between-subjects factor *order of agent form* (robot first or voice assistant first), and the between-subjects factor *morphology priming* (primed or not). Specifically, all groups experienced both forms of Bono—Bot and Voice—one per day over two days. All groups were also randomly assigned to the two between-subjects conditions (n = 16 per condition). Order of agent form meant Bono Bot on the first day and Bono Voice on the second, or vice versa. Morphology priming involved purposely drawing attention to Bono Bot's body or not. At the start of the trial, the primed group received a description about specific features of Bono's morphology (e.g., head shape, facial expression, colours, etc.), while those in the control condition did not. But all groups experienced a manipulation: during the break between sessions on each day when all participants left the room, we added red blush marks to Bono's cheeks.

### 5.2 Participants and Setting

Older adults (14 men and 22 women; 36 total in 9 groups) who had no experience with the Coimagination Method or Bono were recruited using a third-party recruiter from three regions around the city. Each was quasi-randomly assigned to groups of 3–5 people based on their schedules so that they could join the same group on all days. Ages ranged from 65 to 83 (M = 72, SD = 4.4, MD = 72, IQR = 14). All had completed at least a high school education. All had no or mild cognitive impairment and were able to travel to the study location on their own. The Coimagination method required that participants use a smartphone to take and send their own photos to us, which acted as a baseline of technology experience. The study was conducted in a lab space designed for group conversation. Ethics approval was obtained from the RIKEN research ethics committee under No. Wako3 2019-21). Informed consent was obtained and compensation was roughly US $12 per hour plus transportation.

### 5.3 Procedure

The entire study spanned three days for each participant; the study flow is presented in Fig. 4. Participants first joined a training session for the Coimagination Method with a human facilitator (and without Bono) on the first day. At that time, they completed the pre-questionnaire. The main part of the study was then conducted over two days, with a different agent embodiment each day.

Each day followed the same procedure with some exceptions for the robot version. Participants were briefly introduced to the outline for the day. In the priming condition for the robot, participants received a special description about Bono's features (see 5.1). Next was the first of two Coimagination sessions (see 4.2) for the day. On the robot day, Bono's body was modified during the break while participants were outside of the room (see 5.1). The second Coimagination session of the day then commenced. Afterwards, Bono (either form) was hidden by the tech staff. Then participants filled out the agent embodiment questionnaire. After this, a retrospective think-aloud (RTA) session was conducted using a video replay of the trial. RTA, a staple of usability research [77], involves participants speaking their thoughts out loud while watching a video of themselves completing the task [78]. We used a retrospective format because the main task was conversation, making it difficult, if not impossible, to use other approaches. Also, due to the group conversation format, we conducted the RTAs in groups, similar to a focus group but with the structure of an RTA. After the RTA, the day was finished, and participants were compensated. On the last day, participants were debriefed.

After the main study was completed, the PI interviewed the human facilitator, an experienced facilitator of the Coimagination Method who ran the training sessions on the first day and was familiar with Bono Bot. A semi-structured interview procedure was used, starting with the question "In your opinion, when might Bono be better than a human facilitator?" and moving into dynamic and open questions related to facilitation and older adults' reactions to the agent's embodiments.

### 5.4 Instruments and Data Collection

A combination of qualitative and quantitative approaches was used, including questionnaires, interviews, speech data, and observations. All transcripts were translated from Japanese by the PI and checked by a native speaker.

#### 5.4.1 Talkativeness Questionnaire

A pre-questionnaire was developed with three self-report items to assess subjective talkativeness. The first item was "I'm basically a talkative person," with a 3-point agreement scale. The two other items asked for comparative ratings using a 5-point amount scale: "Rate the amount of talking you do during a conversation with a good friend" and "Rate the amount of talking you do during a conversation with someone you haven't had much contact with." See Online Resources 1 and 2 for the English and Japanese versions of the questionnaire, respectively.

#### 5.4.2 Agent Embodiment Questionnaire: Voice and Body Versions

The main instrument was the agent embodiment questionnaire. A Bono Voice version was created, identical except for phrasing and exclusion of the awareness-morphology items. Open-ended questions were added to gather more detailed reports of older adults' experiences: how it felt to have an intelligent agent as a facilitator. An open-ended question was added for the manipulation check: whether participants noticed "anything else over the course of the session" about Bono Bot or Voice, asking about both morphologies so as to avoid unintentionally priming participants.

#### 5.4.3 Speech Data

Speech data was recorded by the agent's conversation-parsing AI in a series of spreadsheets comprised of utterances, timestamps (including start and end of each utterance), and speaker data.

#### 5.4.4 Qualitative Data from the RTAs, Interview, and Observations

The RTA sessions and post-hoc interview with the human facilitator were video-recorded and transcribed by the PI. Observation notes of each Coimagination session were recorded by the PI, a trained researcher who attended all sessions. Observations were categorized by group and day, with participants identified by ID only.

### 5.5 Measures

We generated a series of subjective and objective measures, considering both individual and group configurations.

#### 5.5.1 Awareness-Morphology

The subjective self-report measure from the preliminary study was used (see 4.3.1). Also, an objective measure of awareness was calculated based on the Bono Bot body manipulation and whether it was noticed. Responses to the open-ended manipulation check item in the questionnaire were quantitatively coded as correct or not.

### 5.5.2 Talkativeness

Subjective and objective measures of talkativeness were developed. The three subjective talkativeness items from the pre-questionnaire were summed together. Two objective measures of talkativeness were developed using averages of the recorded speech data. One was the cumulative length of time in milliseconds for all utterances. The other was the number of utterances. Group variables for all of these were generated using means of individual scores.

### 5.5.3 Usability and UX

The self-report measures from the preliminary study was used (see 4.3.2).

## 5.6 Data Analysis

The same data analysis procedure as in the preliminary study was followed (see 4.4), with some additions for the new measures and qualitative data. As before, descriptive and inferential statistics were used to evaluate individuals as well as groups. Inferential statistics were used for exploring relationships and evaluating the hypotheses, including t-tests, Pearson correlations, ANOVAs, and Kruskal-Wallis H tests. Age was divided into three subgroups[2] based on the min and max ages of participants: 13 in 65–69 ("young-old"), 11 in 70–74 ("old-old"), and 10 in 75–80 ("oldest-old"). Two participants did not go to all sessions, so their data was excluded from the comparative analyses.

For the qualitative data, theoretical thematic analyses [79] were conducted by two raters using a corpus comprised of response data from the open-ended questionnaire items, observer notes, interview data from the RTAs, and data from the post-hoc human facilitator interview. We took a "realist" position, treating the data as the experiences and realities of and about participants [79]. One rater produced an initial set of codes from the raw data and then systematically revised the codes with another rater until inter-rater agreement was achieved via a Kappa statistic of 0.80 or higher [80]. Codes that did not achieve this were discarded. The final codes were then categorized into themes based on consensus among the raters and guided by the research question (i.e., agent embodiment, agent facilitation of group conversation, etc.).

## 5.7 Experimental Results

Descriptive statistics can be found in Table 2 in Online Resource 6. We begin by describing the subjective and objective results for the awareness-morphology and talkativeness measures. We then report on the hypothesis results.

### 5.7.1 Awareness-Morphology

The majority (72%) of participants selected the correct answers for the awareness-morphology items. One-quarter had very low scores (zero or one answer correct). However, only one-quarter (26%) correctly noticed the change to Bono Bot's cheeks. Moreover, two-fifths (40%) in the Bono Bot condition and almost one-quarter (22%) in the Bono Voice condition reported another change *that did not happen*. Examples of the reported changes for Bono Bot included movement, more directed speech (to specific participants or in the conversation), and speed. Examples for Bono Voice included improved intonation, stranger intonation, and the appearance of lights on the speaker. A medium positive correlation was found between confidence and accuracy scores for the awareness-morphology measure, $r(34) = 0.728$, $p < .001$, $R^2 = 0.530$. This suggests that confidence is related to performance at assigning correct features to the robot.

### 5.7.2 Talkativeness

Individuals had an average subjective talkativeness score of 3.9 (SD = 1.6, MD = 4, IQR = 5). The group average was 3.9, the same as for individuals (SD = 0.7, MD = 4, IQR = 1.8). A t-test found a significant difference between the lowest and highest scoring groups, $t(4) = -5.43$, $p = .006$, $d = 12.48$, 95% CI [1.23, 1.77], indicating that there may be effects based on group talkativeness. For objective talkativeness, individuals had a cumulative average of 20.7 utterances (SD = 6.7, MD = 21, IQR = 8), with the length of an average utterance being 191.4 ms on average (SD = 74.6, MD = 201.7, IQR = 82.4). Groups had an average of 144.7 utterances (SD = 50.5, MD = 132, IQR = 160) with an average total length of 1339.9 ms (SD = 518.2, MD = 1308.1, IQR = 1327.3).

### 5.7.3 Hypotheses

We now turn to answering the hypotheses.

*H1: The agent's robot embodiment will have higher usability and UX scores than its voice form.*

A paired t-test showed a significant difference in UX scores between robot (M = 96.6, SD = 16.4, MD = 98, IQR = 38) and voice (M = 80.2, SD = 18.2, MD = 79, IQR = 45.5), $t(33) = 5.07$, $p < .001$, $d = 0.07$, 95% CI [-128.71, 170.35], which confirms the hypothesis that the robot form, Bono

---

[2] See gerontologist standards and inclusive language: https://apastyle.apa.org/style-grammar-guidelines/bias-free-language/age.

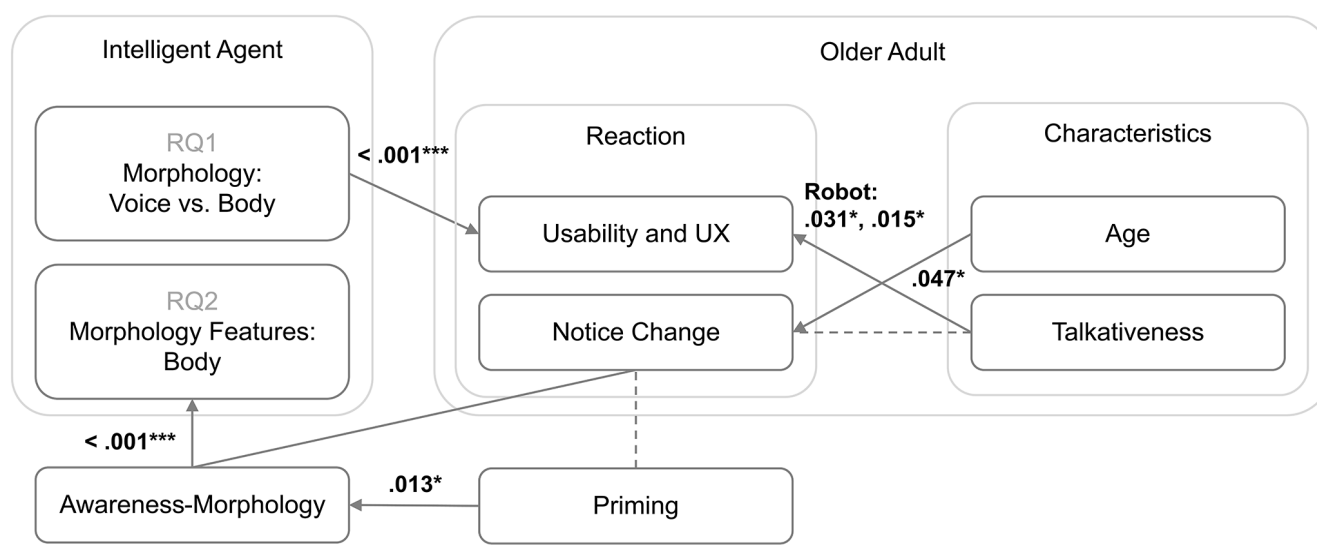

**Fig. 5** Updated hypothetical framework with significant and non-significant results

Bot, was more usable and elicited better UX than the voice form.

*H2: Most participants will notice a change to robot body.*

Since only one-quarter noticed the change, this hypothesis was not confirmed. Even so, a weak correlation was found between awareness-morphology scores and correctly noticing the change, $r(34) = 0.582$, $p < .001$, $R^2 = 0.339$. In effect, those who performed well at the awareness-morphology tasks also tended to notice the change to the robot.

*H3: Priming will increase participants' ability to notice a change to robot body.*

A weak, positive correlation was found between priming and subjective awareness-morphology scores, $r(34) = 0.418$, $p = .014$, $R^2 = 0.175$. A follow-up one-way ANOVA indicated a significant difference, $F(1,32) = 6.755$, $p = .014$, $\eta^2 p = 0.174$, with the primed group (M = 8.27) having higher scores than the control group (M = 4.79). Since the groups were uneven, a Kruskal-Wallis H test was run to confirm these results, and it did, $\chi^2(1) = 6.193$, $p = .013$, $\eta^2 p = 0.158$. However, no significant difference was found for priming based on objective awareness-morphology scores, or the ability to notice the change to the robot's cheeks. Altogether, these results suggest that priming raised the saliency of visual details in the robot's physical form for participants but did not help them to notice when those details were changed. As such, this hypothesis was not confirmed, meaning that priming did not unduly influence the objective measure of awareness-morphology.

*H4: Age will affect participants' ability to notice a change to robot body.*

A negative, weak correlation was found between age and noticing the change, $r(34) = -0.343$, $p = .047$, $R^2 = 0.117$, confirming this hypothesis and indicating that ability to notice the change worsened with age.

*H5: High talkativeness will lead to lower awareness-morphology scores for the robot version of the agent.*

No significant correlations were found, so we can reject this hypothesis.

*H6: High talkativeness will lead to worse UX scores.*

Weak, negative correlations were found for the robot condition between UX scores and cumulative time, $r(34) = -0.370$, $p = .031$, $R^2 = 0.137$, and UX scores and number of utterances, $r(34) = -0.413$, $p = .015$, $R^2 = 0.171$. However, no significant correlations were found for Bono Voice. This suggests that those who spoke more had poorer UX with the robot but not with the voice, partially confirming the hypothesis and contradicting H1, where the robot form received higher ratings.

These results are summarized in Fig. 5.

### 5.7.4 Summary

Most participants had a better experience with Bono Bot compared to Bono voice. However, the more talkative a participant was, the lower they rated their experience with Bono Bot, but not Bono Voice. Talkativeness did not otherwise influence embodiment measures. The majority (almost three-quarters) remembered high-level aspects of the robot's body (i.e., major colours and its unchanging facial expression). Even so, most did not notice the small but salient change to the robot's cheeks. Those that did performed well on the subjective awareness-morphology measure. Our results also indicate that the ability to notice a change in the body worsens with age. Priming did not affect noticing this change, but it did help older adults retain memory of the robot's morphology. Even so, it does not seem that awareness and memory of Bono's stable and changing features affected UX. Taken together, Bono Bot was best overall in terms of UX, apart from those who were more talkative, for whom Bono Voice was best.

### 5.8 Thematic Findings

Here we present the themes from our combined analyses of the corpus of data gathered about participants. Groups were labeled alphabetically from A-I. In most cases, participant IDs have the format of "Group ID plus Participant ID," e.g., C03. For the RTA data only, we use a "r plus Group ID plus Position" format to mark individuals.

#### 5.8.1 Adapting to Agent Facilitation as a Matter of Time

Older adults encountered an adaptation curve when it came to accepting the agent as a facilitator. Some were hesitant or had negative impressions that were allayed over time [B8, D13, E20, H39, I34, I35]. Some were confused at first, but by the end felt positive about the agent and got used to it [B8, D14, E20, H29, H39, I34-36]. As D14 explained, "*At first, I was confused, but then I got used to [the agent], as if it was a friend.*" H39 felt that the gradual switch in feelings towards the agent was "*mysterious.*" But some older adults' feelings worsened over time. F23, for instance, noted that "*after it stopped me for talking too much, I wasn't sure if I could talk again.*" One [I35] felt that both forms of agent were best in the middle but not at the start or end. This highlights the importance of time when it comes to acceptance of the agent.

#### 5.8.2 Specialties of Agent Facilitation: The Superhuman Factor

Older adults highlighted the "superhuman" benefits of "mechanical" agent facilitation. The agent's ability to strictly keep time emerged as a major positive for most (21 comments). As B8 explained: "*There were no mistakes because everything was done by a machine.*" But the benefits went beyond timekeeping. Older adults found themselves thinking about how much they were speaking, realizing that they were speaking too much or too little— in effect, reflecting on and managing their own behaviour in realtime [A1, D15, E19, F23, F24, G38, H29, H39, I35, I36]. As D15 expressed: "*I recognized myself talking too much [the first time] and changed [how much I talked] the next time.*" Older adults were also more open to this kind of management from an agent compared to a human facilitator. Several noted the absence of feeling rejected like they would by a human [D13, D14, D16, D17, E19, F21, F24, G38, I33, I35-36]. As I36 described, "*If you talk too much, it'll tell you to stop [.] but this isn't as unpleasant [...] compared to a human.*" E17 echoed this sentiment: "*[The agent] can proceed without worrying about [us] (in a good way).*" I35 explained: "*Humans in positions of authority are often thought to 'look down on those below,' but robots are softer.*" At the same time, older adults expressed freedom from having to worry about the emotional implications of their participation on others or a human facilitator. As D14 explained: "*[The agent] can't feel emotions, unlike humans.*" I33 also noted that "*not having to worry about the people in the group*" was a great benefit.

#### 5.8.3 Being Interrupted by an Agent: An Affective Triad

One of the key tasks of the agent was interrupting someone when they were speaking too much or calling upon someone when they were speaking too little. Most groups (8 of 9) had visible reactions to the agent (either form) when it interrupted or prompted speech. These reactions took one of three forms: amusement, surprise, or confusion. Participants laughed individually or as a group [e.g., RI2 alone and Group G together] or expressed amusement: "*How cute!*" [rH1]. Surprise reactions resulted from the agent calling on a particular person when they felt that they were not doing anything wrong. As I36 explained: "*I thought I was speaking a lot, but I was encouraged to speak several times.*" While it was individuals that were pinpointed by Bono, the group took these situations as an opportunity for group discussion. For instance, in Group B:

> rB4: "*Why was rB3 cut off by Bono?*"
> rB1: "*I can't believe we talked that much.*"

Some understood what they needed to do in theory but not in practice: "*I thought, 'What should I say?' I need special training*" [rB3]. Others got the message but did not know why it was important. As I36 explained: "*Some people like to listen but [the agent] forced them to speak quickly.*" In contrast to the previous theme, these reactions show how the "mechanical factor" of agents can be a double-edged sword.

#### 5.8.4 Agent Embodiment as a Work in Progress

Most older adults talked about Bono as a "work in progress" [A4, B6-8, E17, E19, G26, G28]. They accepted its current limitations, provided constructive criticism, and mused about the future of AI in their daily lives. One way that acceptance of Bono's shortcomings played out was with humour. Group F, for instance, laughed good-naturedly at the agent's pronunciation of one participant's name. G38 admitted to doing so every time Bono said anyone's name. In Group I, this discussion took place:

> rI3: "*The technology's still maturing, so I tended to ignore the voice.*"
> rI4: "*I agree.*"
> rI3: "*But I was impressed that it knew my name. Even though the pronunciation was weird.*"

Many suggestions were made about Bono's morphology and the underlying AI's abilities. For instance, D14 suggested advanced conversation parsing abilities, such as acknowledging backchannel feedback as legitimate utterances. F24 suggested that "*it should watch for a good time to break up the conversation.*" AI considered how Bono could handle advanced social dynamics like "*when the person doesn't stop talking.*" B6 wished for Bono to "*detect and understand when we get lost and why,*" referring not only to speech recognition but higher-order cognitive abilities and behaviours in order to "*understand our psychology for the Coimagination Method.*" E19 wondered why we were exploring agents: "*Is there a shortage of human moderators? Or is it better to use a robot that can always react fairly? And robots can be used anytime and anywhere!*" Older adults were oriented to curiosity and future-thinking.

#### 5.8.5 An Empathic Voice is Key—for the Robot, Too

Regardless of Bono's embodiment, older adults focused on voice. They provided critical feedback and points for improvement. Intonation, pronunciation, phrasing, and timing were key points of concern across the board. They also expressed a desire for empathic responses from the agent in terms of its voice and body (26 comments each for Bono Bot and Bono Voice). Most desired more interjections or "aizuchi" (30 comments), such as meaningful comments from the agent, questions from the agent, bridging words (e.g., "by the way"), laughter and jokes, and non-linguistic utterances to show that the agent was listening and supportive. Yet only three (D15, G26, I36) commented on changes to the body (gestures and facial expressions), highlighting the importance of voice for expressing empathy.

#### 5.8.6 Still Yearning for the Elusive Human Factor

Most older adults had a positive experience with both forms of Bono, and even pointed out how agents are better than humans for facilitation. Yet, many expressed a desire for "the human factor" in ways that went beyond agent voice [A4, B6, B8, C12, D13, D14, F22, F24, G25, G26, G28, I36]. G26 explained that the agent "*lacks the same kind of warmth as humans.*" G25 wrote: "*Somehow I just want to talk with someone who is flesh and blood.*" About the voice form, F24 wrote: "*It might be easier to talk with if the agent's shape is humanoid,*" a sentiment echoed by E17: "*a human MC-like visual appearance.*" A4 explained that while functionally excellent, the agent was "*not good at grasping emotions and the hearts of people.*" In contrast, others felt that Bono was already humanlike. U39 wrote: "*I felt like I was talking to a person, for some reason.*" E19 focused on Bono Bot in particular, connecting the feeling of humanlikeness to the robot's morphology: "*[Bono Bot] has eyes and a mouth, so it's like talking to a person.*" Older adults' expectations and the state of the technology seem to have coincided.

### 5.9 Insights from the Perspective of Human Facilitation

Although the role and tasks were the same, the human facilitator felt that there are three main differences between human facilitation and agent facilitation. First, the agent, being automated, can interrupt without hesitation. Humans, however, must determine when it is best to intervene. In her case, she considers the flow of the conversation as well as how to adhere to the standard of facilitation expected by the Coimagination method when agent-controlled: "*I ask myself, 'What Bono would do here? Should I cut in?'*" Second, the human facilitator must take care of the feelings of the people participating, a kind of emotional labour. For her, this means providing empathic responses and intervening in ways that do not suggest a rejection of what participants are saying. She expressed a slight fear of participants "*feeling uncomfortable towards me*" after intervening that she has had to build resilience against. Third, the human facilitator can only do one task at a time, while the agent can do other tasks in the background, more or less at the same time. She did not notice or believe that there were any differences between the robot and the voice assistant.

As a quick reference, we provide our matrix of findings in response to RQ1 and RQ2 across the quantitative and qualitative results for voice and body in Table 1.

## 6 Discussion

We now discuss the findings with respect to answering the question of what role agent embodiment plays when the agent is facilitating conversation for groups of older adults.

### 6.1 Agent-Based Facilitation is Effective, if Odd

Older adults' reactions to agent facilitation were positive and nuanced. While it took many older adults time to adapt to an agent facilitator, as hinted at by other work [16], this occurred for almost everyone. The agent achieved high UX scores in general, but especially for its robot form. The automated and machine-like qualities of agent facilitation—the "mechanical factor" and superhuman abilities of the agent—were unexpectedly well-received. We summarize the benefits and oddities below.

Table 1 Matrix of main findings for RQ1 and RQ2 per factor and using all data and analyses

| Factor | Embodiment | Findings | Implication |
|---|---|---|---|
| Usability and UX | Voice | Received lower quantitative ratings; voice received the majority of comments and feedback | Robot may be superior, but voice must not be overlooked; agent facilitation is acceptable and in some ways superior to human facilitation |
| | Body | Received higher quantitative ratings; comments and feedback focused on cuteness as well as low anthropomorphism | |
| | Both | Acceptance of agent facilitation takes time; mechanical factor and superhuman abilities appropriate for group conversation facilitation; agent perceived as a "work in progress"; desire for more natural intonation and emotional expression in the voice | |
| Awareness-Morphology | Voice | Almost one-quarter reported a change that did not occur | Small details in the body do not matter; asking will elicit guesses; age affects performance |
| | Body | About one-quarter noticed the change and reported the correct change; almost three-quarters selected the correct details; ability to notice and report with accuracy decreased with age | |
| Priming | Voice | n/a | Priming brings attention to body details that may otherwise be overlooked |
| | Body | Details of the body made more salient | |
| Talkative-ness | Voice | No relationship between measures and UX | Voice was appropriate for everyone, while body was less effective for talkative types |
| | Body | Cumulative time and number of utterances indicate poorer UX | |

### 6.1.1 Social Exemptions

As mechanical, non-human, and "in progress" facilitators, Bono was a recipient of *social exemptions*. Unlike previous work (cf. [16]), we found that Bono was given a pass for human social transgressions, especially interrupting. This was a critical difference to human facilitation identified by both the older adults and the human facilitator. Yet, like previous work (e.g., [16]), we found that older adults desired cues, bridging statements, and "aizuchi" to smooth over the experience, suggesting limits to the exemption.

### 6.1.2 Affectlessness for Goodwill

For both the older participants and the human facilitator, Bono allowed for what we might call *affectless goodwill*. Almost everyone perceived that Bono had no emotions, was not biased to any particular person, and was developed for a specific task, i.e., group conversation facilitation—in effect, the agent being goal-directed and without feelings encouraged goodwill. Older adults did desire more empathic responses, but this was at the lower level of speech content. In some cases, they responded to Bono calling them out by not taking it personally or seriously, such as by laughing. But this can also be interpreted as Bono serving as a means of emotional management in the moment [81]. In effect, as a target for humorous reactions, Bono provided stress relief when someone did not know what to do. This shows how the valence of laughter is contextual, not always indicating positive affect [20]. It provides a nuanced contrast to the general move towards emotive agents [82]: specifically, the non-affective nature of agents may be beneficial in certain cases or manage emotion in a different way.

### 6.1.3 Standard Adherence

*Standard adherence* was noted by older adults and the human facilitator alike. Being a machine, Bono was perceived to have a built-in ability to stick to the Coimagination Method, with perhaps the occasional glitch. Since one of our practical goals was standardizing the method, this is a good sign of success.

### 6.1.4 Inhuman Multitasking

The human facilitator highlighted the agent's ability to *multitask* in ways that humans cannot do, especially appropriate for the task and context of use in our case. Yet, older adults seemed to desire some human quality that was still missing—the elusive human factor that is not necessarily tied to humanoid bodies or voices.

#### 6.1.5 Automation, Without Oz

Much work in HRI involves the use of robots with limited capabilities and/or relies on Wizard of Oz to bring the robot to life [23, 83]. In contrast, Bono was a fully autonomous facilitator with a working AI. Yet, Bono could not respond to unexpected situations or explain itself. Such situations can lead to an "expectation mismatch" [84, 85] in the case of humanoid robots, especially those that otherwise function well. Two older adults [G25, I35] wondered if Bono had in fact been controlled by a human (or "Wizard"). But the majority considered Bono a "work in progress," musing about how long it might take for it to understand the intricacies of human conversation. This shows how increasing levels of automation can shift, if not improve, perceptions of agent's abilities.

### 6.2 Voice over Body? The Role of Conversation Personality

The influence of agent morphology was a bit hard to tease out, as found previously (e.g., [16]). The robot was better received than the voice assistant in terms of overall UX. Spontaneous discussions about the robot's physical form occurred in the RTA sessions as well as the main sessions. Several older adults took photos with the robot after the study was over. Many commented during and after the study about how "cute" Bono Bot was. Yet, the morphology-awareness scores and qualitative findings suggest that the robot's body was only notable at a high level. Counter to what we expected, the emotional expressiveness of Bono Bot's unchanging sleepy grin was not always memorable [50] and its bodily movements did not necessarily aid in recognition of such visual details in its morphology [53]. This might be explained by the phenomenon of gestalt. In any case, we could not isolate whether specific visual features mattered.

There was one exception when comparing voice and body at a high level: talkative participants, whose experiences with Bono Bot were worse than the rest. This is captured by F23, who had the most utterances out of every participant, and noted: *"For some reason, unlike [with the robot], I had a better impression with [the voice assistant]."* This may be understood if we consider the connection between perceptions of an agent's suitability for a task and user satisfaction measures, as found in previous work [7, 86]. In short, the voice assistant form, which relies on the sound medium, may be better for talkative older adults during group conversation. This contrasts with previous work suggesting that the voice alone may be insufficient (e.g., [49]). Yet, the voice received greater focus than the body, even in the robot condition. This cannot be explained by order effects given the counterbalanced design we used, and we did not reveal the agent forms in advance. Reactions to the voice also do not indicate a clear reason, except perhaps through their sheer variety. At the least, we can conclude that agent morphology was not a factor in terms of the conversation method.

### 6.3 Group Conversation for Grappling with Agent Experiences

The group conversation format allowed participants to share reactions, discuss unexpected situations, and work together to tease out their understanding of the agent and its facilitation. This reflects similar work on older adults' exposure to new technologies (e.g., [87]), with older adults relying on each other to make sense of a new technology and their interactions with it (e.g., [85]). We echo the sentiment that human social bonds in these situations are key. These social sense-making activities, often undergirded with laughter, may have been essential for facilitating the flow of the experience and ensuring a higher level of UX with the agent than a solo activity. These sense-making activities encompassed the entirety of the agent's embodiment—except the physical morphology of Bono Voice, i.e., the speaker, about which we have no record of discussion. A focus was on the context of use and the agent's interactions, especially interruptions by the agent that were unexpected or difficult to understand by individual members. In effect, the group format and conversation setting acted as a space where older adults could grapple with and settle issues that dynamically arose with the agent and its embodiment that they may not have been able to do alone.

### 6.4 Agents Offload Facilitation Workload and Emotional Labour

We have started to broaden the scope of who we should consider when it comes to the design and study of technologies. For instance, Dixon and Lazar [88] highlighted the role of emotional impact and emotional distancing through their interview work with care providers. Similarly, Rea, Geiskkovitch, and Young [89] revealed how researchers and assistants taking on the role of "Wizard" in Wizard of Oz setups can experience harm, even when the normal checks and balances are in place, e.g., ethics procedures. The overarching thread is that we often overlook how our own research can affect ourselves and stakeholders who are not the target users. We found echoes of these sentiments in our interview with the human facilitator, who alluded to the emotional burdens involved in facilitation work, especially in a group setting and when positioned in comparison to an agent. Put simply, an agent does not—indeed, cannot—worry or feel strain in social situations. While a human may be flexible and add experiential value, an agent can offload emotional labour while performing the same tasks as a human. Moreover, the agent can take on other work that a human cannot do; in the case of the Coimagination Method, it can record (as well as respond to) the conversation. In these ways, the agent's lack of humanness can be an asset rather than a detriment.

## 6.5 Implications for Embodiment in Agent-Facilitated Group Conversation with Older Adults

We now offer considerations for the design of similar systems.

### 6.5.1 Focus on Voice, but Do Not Overlook the Body

Voice emerged as a key factor in older adults' experiences across the corpus of data and materials we analyzed and regardless of agent form. Indeed, one clear finding was that the experience of talkative types was worse with the robot form. This provides some support for the move towards voice assistants and other "bodyless" VUIs for older adults, group conversation, and group facilitation. Even in the case of a robot morphology, voice was an important and notable factor in older adults' experiences. Considering the benefits from a practical standpoint, voice-based agents are easier and cheaper to develop, platform-independent, and more portable. Yet, our findings also show that people had a better experience with the robotic form of Bono. The details of Bono's physical form did not seem to be important, however. More work is needed to understand how the mere presence of a body is significant. Regardless, our findings indicate that voice should be given special care—even more than the body.

### 6.5.2 Level Up the Voice: Social and Empathic Utterances

A basic synthetic agent voice may be sufficient and even satisfactory for older adults, but it may also stoke a desire for the elusive human factor. Older adults desire a more humanlike and expressive voice, even in a facilitator. Affective dimensions seem key. What is said also matters. Interjections, including words but also expressive sounds, bridging words, like "So," and context-sensitive timing should be used in agent-led group conversation facilitation. Older adults want to be heard by the agent and know that they are heard, a form of feedback that can be expressed by the agent through aural channels. A group conversation situation can provide the social scaffolds for older adults to navigate their understanding of and experience with intelligent agent facilitators, but this should be supplementary. An agent should be able to rely on its own voice to build rapport and facilitate a positive and understandable experience.

### 6.5.3 Keep It Real with the Superhuman Factor

While they desired more humanlike and expressive voices, older adults also appreciated the mechanical aspects of agent facilitation. This may seem like a contradiction given the binary view underlying a lot of work on the tension between humanlikeness and artificiality in intelligent agents. Yet, as Welge and Hassenzahl [90] have persuasively argued, these mechanical aspects can also act as "superpowers" unique to robots and other agents. Of the six superpowers they outline, "not taking things personally" is especially pertinent for the case of Bono. We can also consider the plurality of embodiment—how each aspect of the agent's morphology, situatedness, and interactions with people do not necessarily point to one answer, even within a single activity. We can mix and match as needed. A humanlike, expressive voice in an agent facilitator may be well-received by older adults, even when its actions are mechanical (e.g., interrupting without social awareness) or superhuman (e.g., processing conversations at a beyond-human clip). Older adults may forgive an agent facilitator that is strict or commits a social faux pas, as long as the agent is perceived as not human in some respect. Future work will need to tease out when and under what conditions humanlikeness and mechanicalness can be fruitfully mixed.

### 6.5.4 Satisfy an Appetite for Co-Creation through Human-Agent Co-Facilitation

The tension between the mechanical and human factors suggests a new approach: bringing together agent and human facilitation in one system. The agent could take on the technically and socially difficult aspects of facilitation—especially time-keeping and interrupting people—while the human facilitator could maintain the mood, manage unexpected situations, and provide the sought-after sense of a human presence. This approach is not unprecedented, with previous work showing that older adults desire human intervention when robot automation is insufficient (e.g., [85]). Designing such a co-facilitation platform will need to be done with human facilitators. As our interview findings suggest, special care will need to be given to the psychosocial burdens that human facilitators may experience. Following other work in this area (e.g., [88]), a first step could be participatory design workshops with older adults, human facilitators, and technologists and/or interviews with the human facilitators. These could focus on developing use cases and scenarios as well as user needs for providing and receiving co-facilitation across the suite of potential users and stakeholders.

## 6.6 Limitations and Future Work

The number of groups was relatively small, and we interviewed only one human facilitator. We recognize that our findings can be bolstered with the addition of more participants. This project, like many others [91], was affected by the COVID-19 pandemic. Due to our research design, we could not move the study online. Instead, we added the follow-up human facilitator interview to gather similar, albeit not equivalent insights. We also underutilized the speech data recorded by the AI, i.e., only for talkativeness. We plan to explore how to use the data to quantitatively measure the effect of agent interventions, e.g., speaking more or less. We also did not measure common HRI factors, such as trust, so as to keep the length of the questionnaire and sessions manageable for older adults. We plan to do so in future work. Finally, the current study took place in an institutional setting with people who were strangers. Future work can consider community centres, hospitals, and care homes. As others have shown, technologies that become domesticated [92] through everyday situatedness as well as those that are participatory may bring about beneficial effects. Future work could, for instance, explore co-designing Bono's features with older adults.

## 7 Conclusion

Intelligent agent-facilitated group conversation, like Bono's Coimagination Method, has great potential for scalability and suitability in older adults' lives. What form the agent should take is harder to pin down. Voice appears to be key for bodied and bodiless morphologies. Trade-offs may be needed for other aspects of agent embodiment and in particular morphology. These may be specific to the activity performed by the agent and its role (i.e., control of group conversation) or the personality of participants in group settings (i.e., talkativeness). Indeed, the voice assistant was less disruptive for more talkative types. Regardless of morphology, the superhuman factor of agent facilitation appears to be beneficial in group conversation contexts. Despite this, older adults still desired a greater sense of the "human factor." Involving human co-facilitators as well as focusing on the voice of the agent may be the way forward.

**Acknowledgements** This work was funded by Grants-in-Aid for Scientific Research (18KT0035; 19H01138; 20H05022; 20H05574) and AMED (JP19he2002014). We thank Kaai Yamaguchi, Kazuhiro Tamura, and Hiroyuki Taira for research assistance. We thank the Fonobono Institute and the Silver Human Resources Center for recruitment assistance. Our sincere gratitude to Peter Pennefather, Jacqueline Urakami, and Yvonne Rogers for reviewing earlier drafts of this manuscript.

**Data Availability** We can provide anonymized and summarized datasets on reasonable request.

## Declarations